\documentclass{article}

\usepackage{arxiv}

\usepackage[utf8]{inputenc} 
\usepackage[T1]{fontenc}    
\usepackage{hyperref}       
\usepackage{url}            
\usepackage{booktabs}       
\usepackage{amsfonts}       
\usepackage{nicefrac}       
\usepackage{microtype}      
\usepackage{lipsum}

\usepackage{amsmath,amssymb} 

\usepackage{paralist}
\usepackage[inline,shortlabels]{enumitem}
\usepackage{multirow}
\usepackage{adjustbox}

\usepackage{array}
\newcolumntype{C}[1]{>{\centering\arraybackslash}p{#1}}
\newcolumntype{P}[1]{>{\centering\arraybackslash}p{#1}}

\usepackage{algorithm, algorithmic}
\usepackage{booktabs}

\usepackage{diagbox}

\title{Keyword Spotting Simplified: \\ 
A Segmentation-Free Approach using \\
Character Counting and CTC re-scoring}
\author{
  George Retsinas\\
  School of Electrical \& Computer Engineering\\
  National Technical University of Athens, Greece\\
  \texttt{gretsinas@central.ntua.gr} \\
   \And
 Giorgos Sfikas\\
 Department of Surveying \& Geoinformatics Engineering\\ 
University of West Attica, Greece\\
  \texttt{gsfikas@uniwa.gr} \\
  \And
 Christophoros Nikou \\
  Department of Computer Science and Engineering\\
University of Ioannina, Greece\\
  \texttt{cnikou@cse.uoi.gr} \\
}

\begin{document}

\maketitle
\begin{abstract}
Recent advances in segmentation-free keyword spotting borrow from state-of-the-art object detection systems to simultaneously propose a word bounding box proposal mechanism and compute a corresponding representation.
Contrary to the norm of such methods that rely on complex and large DNN models, 
we propose a novel segmentation-free system that efficiently scans a document image to find rectangular areas that include the query information. 
The underlying model is simple and compact, predicting character occurrences over rectangular areas through an implicitly learned scale map, trained on word-level annotated images. 
The proposed document scanning is then performed using this character counting in a cost-effective manner via integral images and binary search. 
Finally, the retrieval similarity by character counting is refined by a pyramidal representation and a CTC-based re-scoring algorithm, fully utilizing the trained CNN model. 
Experimental validation on two widely-used datasets shows that our method achieves state-of-the-art results outperforming the more complex alternatives, despite the simplicity of the underlying model.

\end{abstract}

\section{Introduction}

Keyword spotting (or simply \emph{word} spotting) has emerged as an alternative to handwritten text recognition, providing a practical tool for efficient indexing and searching in document analysis systems. 
Contrary to full handwriting text recognition (HTR), where a character decoding output is the goal, spotting approaches typically involve a soft-selection step, 
enabling them to ``recover'' words that could have been potentially been assigned a more or less erroneous decoding.
Keyword spotting is closely related to text localization and recognition.
The latter problems are typically cast in an ``in-the-wild'' setting, where we are not dealing with a document image, but rather a natural image / photo that may contain patches of text.
The processing pipeline involves producing a set of candidate bounding boxes that contain text and afterwards performing text recognition~\cite{jaderberg2014deep} or localization and recognition are  both part of a multi-task loss function \cite{lyu2018mask,zhang2022mask}.
The main difference between methods geared for in-the-wild detection and localization and document-oriented methods is that 
we must expect numerous instances of the same character on the latter; 
differences in character appearance in the wild relate to perspective distortions and font diversity, while in documents perspective distortion is expected to be minimal;
also, information is much more structured in documents, which opens the possibility to encode this prior knowledge in some form of model inductive bias.

In more detail, keyword spotting (KWS) systems can be categorized w.r.t. a number of different taxonomies.
Depending on the type of query used, we have either Query-by-Example (QbE) or Query-by-String (QbS) spotting.
The two settings correspond to using a word image query or a string query respectively.
Another taxonomy of KWS systems involves categorization into segmentation-based and segmentation-free systems.
These differ w.r.t. whether we can assume that the document collection that we are searching has been pre-segmented into search target tokens or not;
the search targets will usually be word images, containing a single word, or line images, containing a set of words residing in a single text line.
Segmentation-based methods involve a simpler task than segmentation-free methods, but in practice correctly segmenting a document into
words and lines can be a very non-trivial task, especially in the context of documents that involve a highly complex structure such as tabular data,
or manuscripts that include an abundance of marginalia, etc.

The former category motivated a representation-driven line of research, 
with Pyramidal Histogram of Characters (PHOC) embeddings being the most notable example of attribute-based representation \cite{almazan2013handwritten, sudholt2016phocnet, sudholt2017evaluating, sfikas2017phoc, retsinas2018exploring, retsinas2018compact, krishnan2018hwnet, retsinas2022fly}.
Sharing the same main concept, a different embedding was introduced by Wilkinson et al.~\cite{wilkinson2016semantic}, while recognition-based systems were also used to tackle the problem in a representation level ~\cite{krishnan2018word, retsinas2021from}.
Contrary to segmentation-based literature,  modern systems capable of segmentation-free retrieval on handwritten documents are limited. 
A commonly-used methodology is the straightforward sliding window approach,  
with \cite{rothacker2013bag, ghosh2015query, ghosh2015sliding} being notable examples of efficient variations of this concept.
Line-level segmentation-free detection uses essentially a similar concept in its core~\cite{retsinas2019alternative}, whilst simplified due to the sequential nature of text-line processing.
Another major direction is the generation of word region candidates before applying a segmentation-based ranking approach (e.g. using PHOC representations)~\cite{rothacker2017word, ghosh2017r, wilkinson2017neural, zhao2020query}
Candidate region proposal can be part of an end-to-end architecture, following the state-of-the-art object detection literature, as in \cite{wilkinson2017neural, zhao2020query},  at  the cost of a document-level training procedure that may require generation of synthetic data.


In this work, we focus on the segmentation-free setting, where no prior information over the word location is known over a document page. 
Specifically, we aim to bypass any segmentation step and provide a spotting method that works in document images without sacrificing efficiency. 
The main idea of this work is to effectively utilize a per-pixel character existence estimate.
To this end, we first build a CNN-based system that transforms the document input image into a down-scaled map of potential characters.
Then, by casting KWS as a \textbf{character counting} problem, we aim to find bounding boxes that contain the requested characters.
In a nutshell, our contributions can be summarized as follows.\\
\begin{itemize}
    \item 
    We build a computationally efficient segmentation-free KWS system; 
    \item 
    We propose a training scheme for computing a character counting map and consequently a counting model; training is performed on segmented words, with no need for document-level annotation 
    which may require synthetic generation of images to capture large variations over the localization of word images~\cite{wilkinson2017neural}.  
    \item 
   We introduce a document scanning approach for the counting problem with several computational-improving modifications, 
    including integral images of one-step sum computations for counting characters 
    and binary search for efficient detection of bounding boxes.
    \item 
    We further improve the spotting performance of the counting system; 
    a finer KWS approach is used on the subset of detected bounding boxes to enhance performance and reduce counting ambiguities 
    (i.e., ``end'' vs ``den''). 
    \item
    In this work, we use the trained CNN model in two variations: a pyramidal representation of counting, akin to PHOC embeddings \cite{Almazan14PAMI}
    and a CTC-based scoring approach, akin to forced alignment \cite{toselli2016hmm}.
    The latter method is also capable of enhancing the detection of the bounding box, notably improving performance. 
    \item 
    We propose a new metric to quantify bounding box overlap, replacing the standard Intersection over Union (IoU) metric.
    We argue that the new metric is more suitable than IoU in the context of KWS, as it does not penalize enlarged detected boxes 
    that do no contain any misdetected word. 
\end{itemize}

The effectiveness of the proposed method serves as a counter-argument to using object detection methods with complex prediction heads and a predefined maximum number of detections (e.g. \cite{wilkinson2017neural}) and showcase that state-of-the-art results can be attained with an intuitively simple pipeline.  
Moreover, contrary to the majority of existing approaches, the proposed method enables sub-word or multi-word search since there is no restriction over the bounding box prediction. 

\newpage

\section{Proposed Methodology}


Our core idea can be summarized as: 
``\textit{efficiently estimate the bounding box by counting character occurrences}". In this section, we will describe how 1) to build and train such a character counting network, 2) use character counts to efficiently estimate a bounding box, and 3) provide an enhanced similarity score to boost performance.
We will only explore the QbS paradigm, which is in line with a character-level prediction system, however extending this into a simple QbE system is straightforward (e.g., estimate character occurrences into an example image and use this as target).

\subsection{Training of Character Counting Network}
\vspace{-.1cm}
\subsubsection{Method formulation:}
The simplest way to treat the counting problem is through regression: given a per-character histogram of character occurrences, one can train a regression DNN with mean squared loss that takes as input word images. 
Nonetheless, such an approach lacks the ability to be easily applicable to page-level images. 
A key concept in this work is the per-pixel analysis of the character probabilities and their scale. 
This way, the desired counting operation can be decoupled into two sub-problems:
1) compute the character probability at each point of the feature map and 2) compute the scale corresponding to each point (i.e., the size of the point w.r.t. the whole character it belongs to), such that summation over the word gives us the requested counting histogram.

Formally, we denote as $F$ the feature map that contains character-level predictions of their probability, and is generated by a deep neural network. This 3D tensor $F$ has size equal to $H_r \times W_r \times C$, where $H_r \times W_r$ is the downscaled size of the initial $H\times W$ size of the input image and $C$ is the number of possible characters. 
We also generate a character-independent scale matrix $S$, also from a DNN, of size $H_r \times W_r$. 
Then the scaled feature map is denoted as $F_s$, where each spatial point of $F$ is multiplied by the corresponding scale value, that is $F_s[i,j,k] = F[i,j,k] \cdot S[i,j]$.
Given a bounding box that contains a word, denoted by the starting point $(s_i, s_j)$ and the ending point $(e_i, e_j)$, one can compute the character occurrences $y_c$ as:
\vspace{-.2cm}
\begin{equation}
y_c = \sum_{i=s_i}^{e_i} \sum_{j=s_j}^{e_j} F_s(i,j)\, , \,\, y_c \in \mathbb{R}^{C} 
\label{eq:cpred}
\vspace{-.1cm}
\end{equation}
The latter formulation can be used to straightforwardly regress the models with respect to the target count histogram. 

To assist this decoupling approach, we also constrain the feature map $F$ to be in line with a handwritten text recognition system, using a CTC loss~\cite{graves2006connectionist}. 
In other words, character probabilities are explicitly trained with this extra loss. 
Note that, even though we do not explicitly train the scale map $S$, it is actually implicitly learned to correspond to scale-like predictions through the elegant combination of the CTC and counting regression loss.

\begin{figure}[b]
    \centering
    \includegraphics[width=\linewidth]{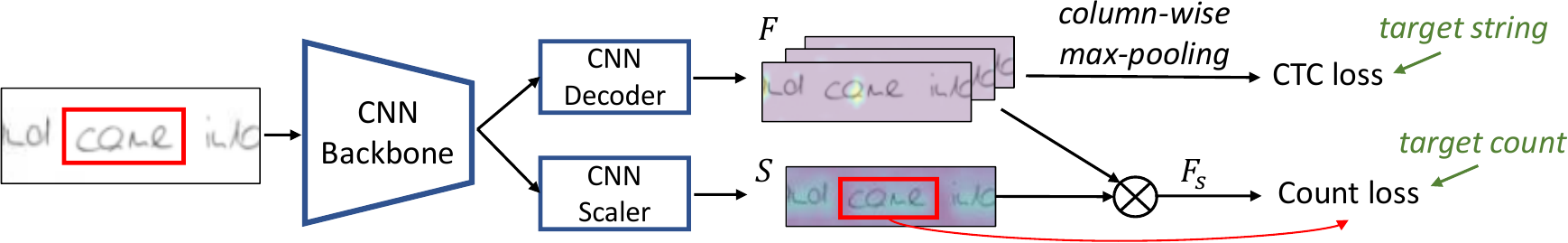}
    \caption{Overview of the model's components and how they contribute to the losses.}
    \label{fig:train}
\end{figure}

The previous formulation is depicted in Figure~\ref{fig:train}. We first feed an image to a \textit{CNN backbone} and obtain a feature tensor, which is then used by the \textit{CNN Decoder} to predict the feature map $F$, and also by the \textit{CNN Scaler}, to predict the scale map $S$. Then, the feature map is multiplied with the scale map $S$, and used to calculate the counting regression loss. The feature map is also independently used to calculate the CTC loss. 


\subsubsection{Model's Components:} 

Having described now the main idea, we outline in more detail the architecture of each CNN component. Note that we did not thoroughly explore these architectures, since it is not the core goal of our paper. In fact, since efficiency is the main point of this approach, we used a lightweight ResNet-like network\cite{retsinas2022best}, consisted of the following components:\\
    \vspace{-.1cm}\\
    $\bullet$ \emph{CNN Backbone:}
    The CNN backbone is built by stacking multiple residual blocks\cite{he2016deep}. The first layer is a $7 \times 7$ convolution of stride 2 with $32$ output channels, followed by cascades of $3 \times 3$ ResNet blocks (2 blocks of 64, 4 blocks of 128 and 4 blocks of 256 output channels). All convolutions are followed by ReLU and batch-norm layers. Between cascades of blocks max-pooling downscaling of $\frac{1}{2}$ is applied, resulting to an overall downscale of $\frac{1}{8}$.\\
    \vspace{-.1cm}\\
    $\bullet$ \emph{CNN Decoder:} 
    The CNN Decoder consists of two layers; one is a simple $3\times 3$ conv. operation of 128 output channels, and it is followed by an $1\times 5$ convolution, since we assume horizontal writing. The latter conv. operation has $C$ output channels, as many as the characters to be predicted.
    Between the two layers, we added ReLU, batch-norm, and Dropout.
    Note that we use only conv. operations so that the decoder can be applied to whole pages. If we used recurrent alternatives (e.g. LSTM), a sequential order should be defined, which is not feasible efficiently considering the 2D structure of a raw document page.  \\
    \vspace{-.1cm}\\
    $\bullet$ \emph{CNN Scaler:}
    The CNN Scaler also consists of two conv. layers, both with kernel of size $3 \times 3$. The first has 128 output channels and the second, as expected, only one, i.e. the scale value. Again, between the two conv. layers, we added ReLU, batch-norm and Dropout.
    Finally, we apply a sigmoid function over the output to constrain the range of the scale between 0 and 1 (when the scale $s=1$ then the pixel corresponds to a whole character).

These components and their functionality are visualized in Figure~\ref{fig:train}.
As we highlighted earlier, the aforementioned architecture contains a novel and crucial modification; it includes a \textit{separate} scale map that enables character counting and is trained \textit{implicitly} as a auxiliary path.  

\vspace{-.2cm}
\subsubsection{Training Details/Extra modifications:}

\begin{figure}[t]
    \centering
    \includegraphics[width=.8\linewidth]{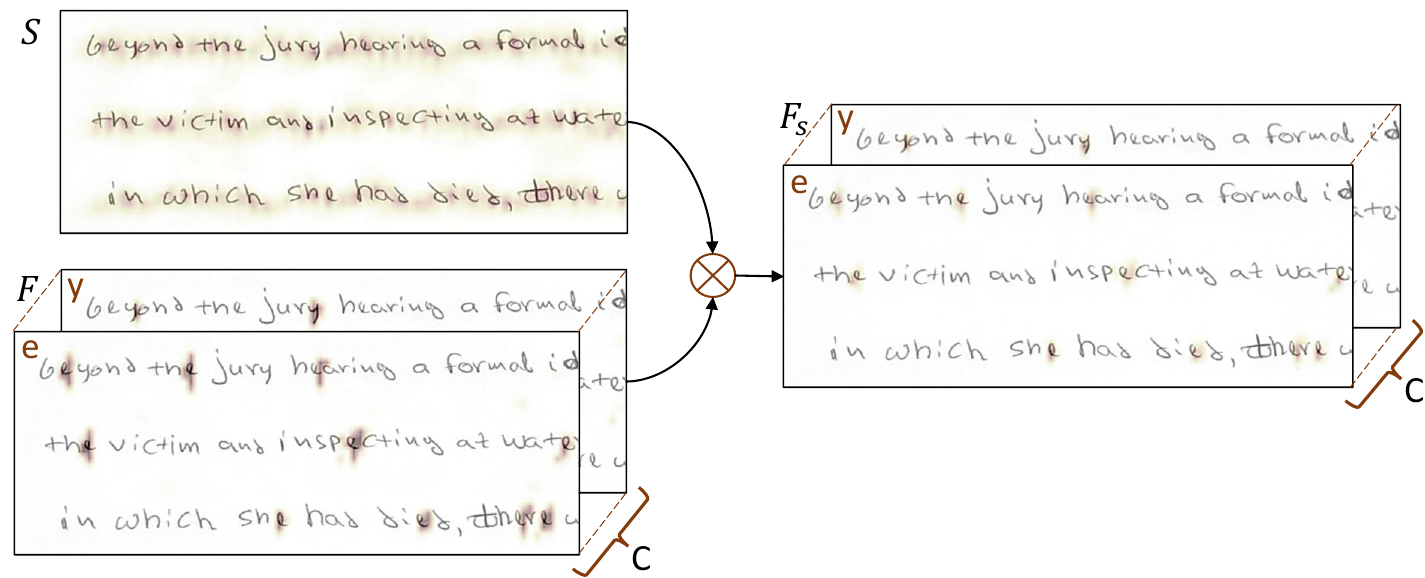}
    \vspace{-.1cm}
    \caption{Document-level extraction of feature maps. The scale map $S$ is multiplied with character probability map $F$ to generate $F_s$. For the 3D tensors $F$, $F_s$ of $C$ channels, we indicatively show the per-character activation of characters `e' and `y'.}
    \label{fig:examples}
    \vspace{-.4cm}
\end{figure}

The feature map $F$ of character probabilities is essentially the output of the CNN decoder, after applying a per-pixel softmax operation over the characters' dimension.
This way, we can straightforwardly generate page-level detections of characters as seen in Figure~\ref{fig:examples}.
Nonetheless, training using the CTC loss requires a sequence of predictions and not a 2D feature map.
To this end, during word-level training, we apply a flattening operation before the aforementioned softmax operation.
Specifically, we use the column-wise max-pooling used in~\cite{retsinas2022best}.

We also used the following modifications to assist training: 1) we extract a larger region around a word to learn that different characters may exist outside the bounding box as neighboring ``noise" and 2) we penalize pixels outside the bounding box before the column-wise maxpooling operation and thus force them to be ignored during the recognition step (we use constant negative values since we have a max operation)  

Overall, the loss used for training was $L_{CTC} + 10 \cdot L_{count}$, where: \textbf{1)} $L_{CTC} = CTC(F_{max}, s_{target})$ is the recognition loss, with $F_{max}$ being the column-wise max-pooled version of $F$ with penalized values outside of the bounding box and $s_{target}$ the target word text, \emph{and}
\textbf{2)} $L_{count} = ||y_c - t_c||_2$ is the counting regression loss, where $y_c$ is the predicted counting histogram calculated as in Eq.~\ref{eq:cpred} and $t_c$ the target counting histogram. 
The weights of each individual loss were set empirically with no further exploration. 
Since the regression loss is more important for our case, it was assigned a larger weight.

\vspace{-.2cm}
\subsection{Efficient Spotting using Character Counting}
\label{sec:proposed-spotting}
\vspace{-.1cm}

Having defined the counting architecture and its training procedure, we proceed with our major goal: 
utilize the generated counting map to efficiently detect bounding boxes of queries.
To this end, in what follows, we will present different sub-modules designed to provide fast and accurate predictions.
Figure~\ref{fig:overview} contains a visualization of the proposed pipeline.

\vspace{-.3cm}
\subsubsection{Problem Statement \& Initial Complexity:}
Here, the input is the per-character scaled map $F_s$, as generated by the proposed architecture. 
The task is to \textit{estimate a bounding box that contains a character count that is similar to the query}.
Note that this way, the detection is size-invariant - the bounding box can be arbitrarily large.
The simplest way to perform such actions is extremely ineffective; 
for each pixel of the (downscaled) feature map, one should check all the possible bounding boxes and compute their sum, 
resulting to an impractical complexity of $\mathcal{O}(N_r^3)$, where $N_r  =  H_r \times W_r$ is the number of pixels  of the feature map.
This can be considered as a naive sliding box approach.
\vspace{-.3cm}
\subsubsection{Cost-Free Summation with Integral Images:}
Since the operation of interest is summation, the use of integral images can decrease this summation step of arbitrary-sized boxes to a constant $\mathcal{O}(1)$ step. The use of integral images is widespread in computer vision applications with SURF features as a notable example \cite{bay2008speeded}, while Ghosh et al. \cite{ghosh2015sliding} used this concept to also speed-up segmentation-free word spotting. Moreover, as we will describe in what follows, the proposed detection algorithm heavily relies on integral images for introducing several efficient modifications. 

\begin{figure}[t]
    \centering
    \includegraphics[width=1.0\linewidth]{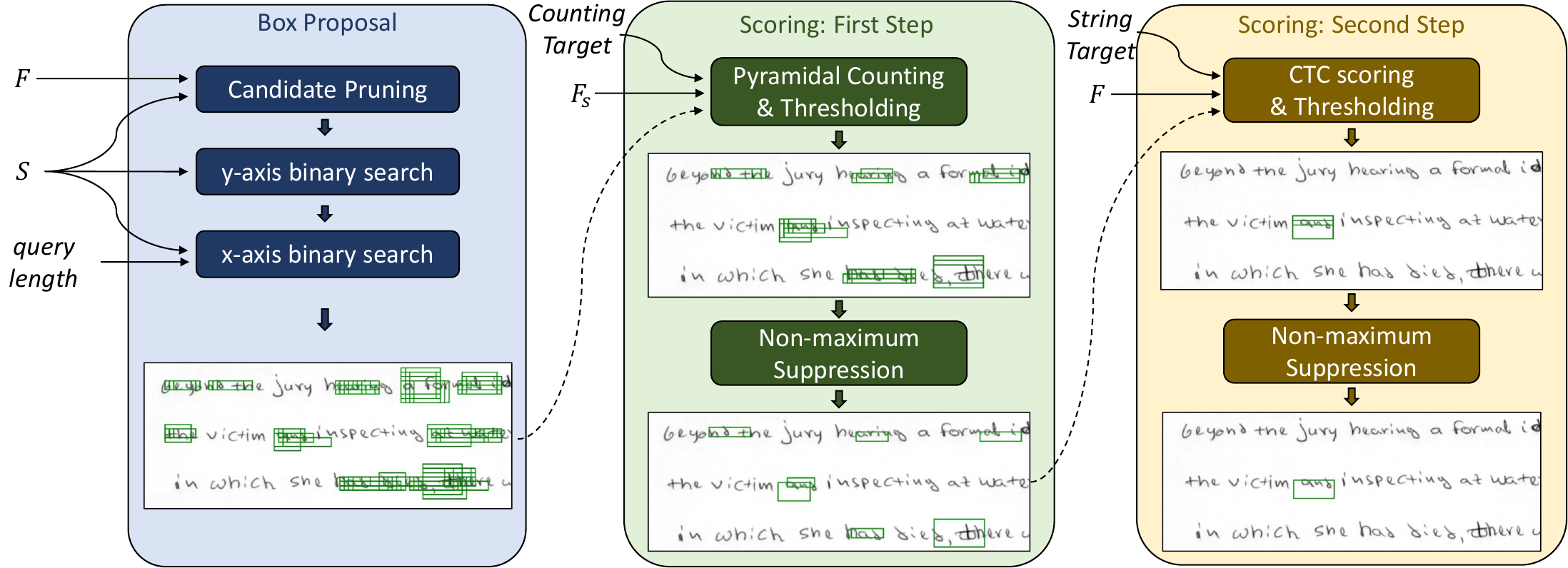}
    \vspace{-.4cm}
    \caption{Overview of the proposed spotting pipeline. The box proposal stage is described in Sec.~\ref{sec:proposed-spotting}, while the scoring steps are described in Sec.~\ref{sec:scoring}.}
    \label{fig:overview}
    \vspace{-.4cm}
\end{figure}

\vspace{-.2cm}
\subsubsection{Bounding Box Estimation with Binary Search:}
Next, we focus on how to search through any window size without the need to actually parse the whole image. 
Specifically consider a starting point $(x_s,y_s)$ on the feature map. 
We seek the ending point $(x_e, y_e)$, where the counting result in the bounding box, defined by these two points, is as close as possible to the character count of the query.
Here, we make use of a simple property of the generated feature map. 
The counting result should be \textbf{increasing} as we go further and further from the starting point. 
This property enables us to break this task into two simple increasing sub-tasks that can be efficiently addressed by \emph{binary search} operations.
Note that we compute the required counting value on-the-fly by using integral images (only the starting point and ending point suffice to compute the requested summation).

\begin{figure}[h]
    \centering
    \begin{tabular}{cc}
     \includegraphics[width=.26\linewidth]{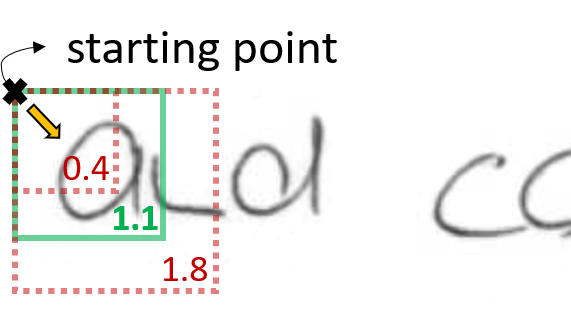}  &  
     \includegraphics[width=.25\linewidth]{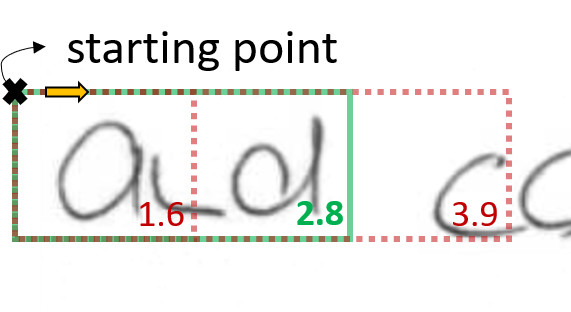} 
     \\
     (a) & (b) \\
     \multicolumn{2}{c}{
     \includegraphics[width=.38\linewidth, trim={0.8cm .1cm 1.8cm .0cm},clip]{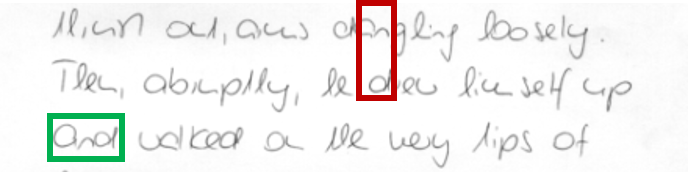}
     } \\
     \multicolumn{2}{c}{(c)}
    \end{tabular}
    \caption{(a,b): Binary search operations over the scale map $S$ to find the height (a) and the width (b) of a possible word. (c): Visualization of the issue of correct counting across lines. Both boxes contain the requested character counting for the query `and'.}
    \label{fig:bsearchs}
\end{figure}

To further simplify the process we act only on the integral image of scale $S$ and we perform the following steps:
    \textbf{1)}
    Find a rectangular area of count equal to $1$. 
    This is the equivalent of detecting one character. 
    We assume that the side of the detected rectangle is a good estimation of a word height on this specific location of the document.  This operation is query independent and can be pre-computed for each image.
    This step is depicted in Figure~\ref{fig:bsearchs}(a).
    \textbf{2)}
    Given the height of the search area, find the width that includes the requested character count. 
    This step is depicted in Figure~\ref{fig:bsearchs}(b).

The aforementioned two-step procedure reasoning is two-fold; first it produces candidate bounding boxes with minor computational requirements and it also resolves possible axis ambiguity. The latter problem can be seen in Figure~\ref{fig:bsearchs}(c), where we can find a ``correct'' bounding box, containing the requested counting histogram, across neighboring text-lines.

\subsubsection{Candidate Point Pruning:}
Up to this point, the proposed modifications notably decreased the algorithm complexity to $\mathcal{O}(N_r\, \log N_r)$.
Nonetheless, traversing through all points is unnecessary and we can further decrease computational requirements via a \emph{pruning} stage over possible starting points. 
We distinguish two useful (heuristic) actions that can considerably reduce the search space:\\
    \textbf{1)}
    The starting point should have (or be close to a point that has) a non-trivial probability over the first character of the query. 
    Implementation-wise, a max-pooling operation with kernel size $3$ (morphological dilation) is used to simulate the proximity property, 
    while a probability threshold is used to discard points. 
    The threshold is relatively low (e.g.,  $0.1$), in order to allow the method to be relatively robust to partial character misclassification.\\
    \textbf{2)}
    Find only ``well-centered'' bounding boxes. 
    This step is implemented through the use of integral images, where we compute the counting sum over a reduced window over y-axis (height) in the center of the initially detected bounding box. 
    If the counting sum of the subregion divided by the sum of the whole word region is lower than a threshold ratio, we discard the point. 
    In other words, we try to validate if the included information in the bounding box corresponds to ``centered'' characters, or it could be a cross-line  summation of character parts.

\subsection{Similarity Scoring}
\label{sec:scoring}

The proposed detection algorithm acts in a character-agnostic way with the exception of the first character of the query. 
Now, we have to find the most similar regions w.r.t. to the actual query. 
The most straightforward solution is to compute the per-character counting inside the bounding box and compare it to the query counting. 
The predicted counting can be computed with minor overhead by using the integral image rationale over the extracted scaled character probability map. 
Comparison is performed via cosine similarity.

\begin{figure}
    \centering
    \begin{tabular}{cc}
    \includegraphics[width=.46\linewidth]{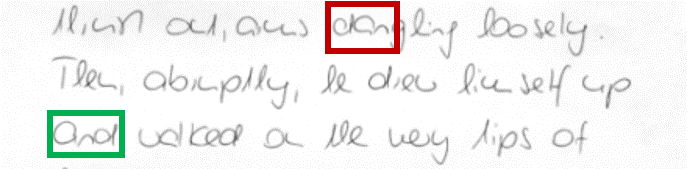}
    &
    \includegraphics[width=.46\linewidth] {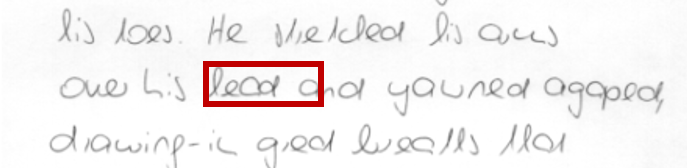}
    \end{tabular}
    \caption{Examples of two possible detection problems: character order ambiguity (left) and box extension to neighboring words (right).}
    \label{fig:ambig}
\end{figure}

A counting-based retrieval cannot distinguish between different permutations of the query characters.
For example, ``and'' and ``dan'' of Figure~\ref{fig:ambig}(left) have the same counting description, confusing the system.
Therefore, the necessity to alleviate ambiguities comes in the limelight.
Towards addressing this problem, we propose two different approaches that can be combined into a single method, as distinct steps, and have a common characteristic: 
the \emph{already trained network} is appropriately utilized to effectively predict more accurate scores. Both steps are followed by a typical non-maximum suppression step.
Figure~\ref{fig:overview} depicts these steps, as well as their outputs, in detail.

\subsubsection{First Step: Pyramidal Counting}
A straightforward extension of the basic counting procedure is building a descriptor in a pyramidal structure, akin to PHOC~\cite{Almazan14PAMI, sudholt2016phocnet}. 
The idea is simple: break the query string into uniform parts and compute the character count into each part. 
This is repeated for $l$ levels, with $l$ counting histograms at each vector, resulting to a descriptor of size $=l (l + 1) C / 2$. 
A Gaussian-based weighting is used for characters belonging to more than one segments, in order to distribute the value accordingly along the segments.

Given the pyramidal query descriptor, one must find its similarity to the respective predicted pyramidal counting over the detected boxes. 
This is also efficiently performed by using integral images over $F_s$ to compute sub-area sums 
of horizontal segments in the interior of each detected box.
Matching is then performed by computing cosine similarity. 
In its simplest form, this first step computes the cosine similarity of the count histograms with only 1 level.

\subsubsection{Second Step: CTC Re-scoring}
For this re-scoring step, we consider a more complex approach based on CTC loss with a higher computational overhead. 
Thus, we use the descriptors of the first step to limit the possible region candidates.
This is performed by a non-maximum suppression step with a low IoU threshold ($=0.2$) that prunes overlapping regions. 
Subsequently, the top $K=30$ results per document page are considered for this re-scoring step. 

Given a detected box, one can compute the CTC score from the already extracted map, resembling forced-alignment rationale. 
Implementation-wise, to avoid a max-pooling operation of different kernels at each region, we pre-computed a vertical max-pooling of kernel $=3$ over the character probabilities map. Subsequently, the sequence of character probabilities was extracted from the centered $y$-value of the bounding box under consideration.

Despite the intuitive concept this approach, as is, under-performed, often resulting to worse results compared to a pyramidal counting with many levels. 
A common occurring problem, responsible for this unexpected performance inferiority, was the inaccurate estimation of bounding boxes that extended to neighboring words, as seen in Figure~\ref{fig:ambig}(right). 
Such erroneous predictions are typically found if a character existing in the query also appears in neighboring words.
Note that the used counting approach cannot provide extremely accurate detections and we do not expect it to do so. 

The solution to this issue is simple when considering the CTC algorithm: 
a score matrix $D \in \mathbb{R}^{T \times C}$ emerges, where the score $D[t,c]$ corresponds to step $t = 0, \dots, T-1$ 
assuming that the character $c = 0,\dots, C-1$ exists at this step. 
Therefore, instead of selecting the score of the last query character 
(alongside the blank character - but we omit this part for simplicity) at the last step, we search for the best score of the last character over all steps, meaning that the whole query sequence should be recognized and only redundant predictions are omitted.
To assist this approach, an overestimation of the end point is considered over the $x$-axis. 
This approach is straightforwardly extended to correct the starting point of the box, by inverting the sequence of probabilities along with the query characters. 

Summing this procedure up, we can use the recognition-based CTC algorithm over the detected box to not only provide \emph{improved scores}, but also \emph{enhance the bounding box prediction}. 
As one can deduct, the re-scoring step can be performed by an independent (large) model to further increase performance, if needed. 
In this work, however, it is of great interest to fully utilize the already existing model and simultaneously avoid adding an 
extra deep learning component which may add considerable computational overhead.

\section{Revisiting Overlap Metrics}
\label{sec:iou}

\begin{figure}[h]
    \centering
    \begin{tabular}{cc}
      \includegraphics[width=.40\linewidth]{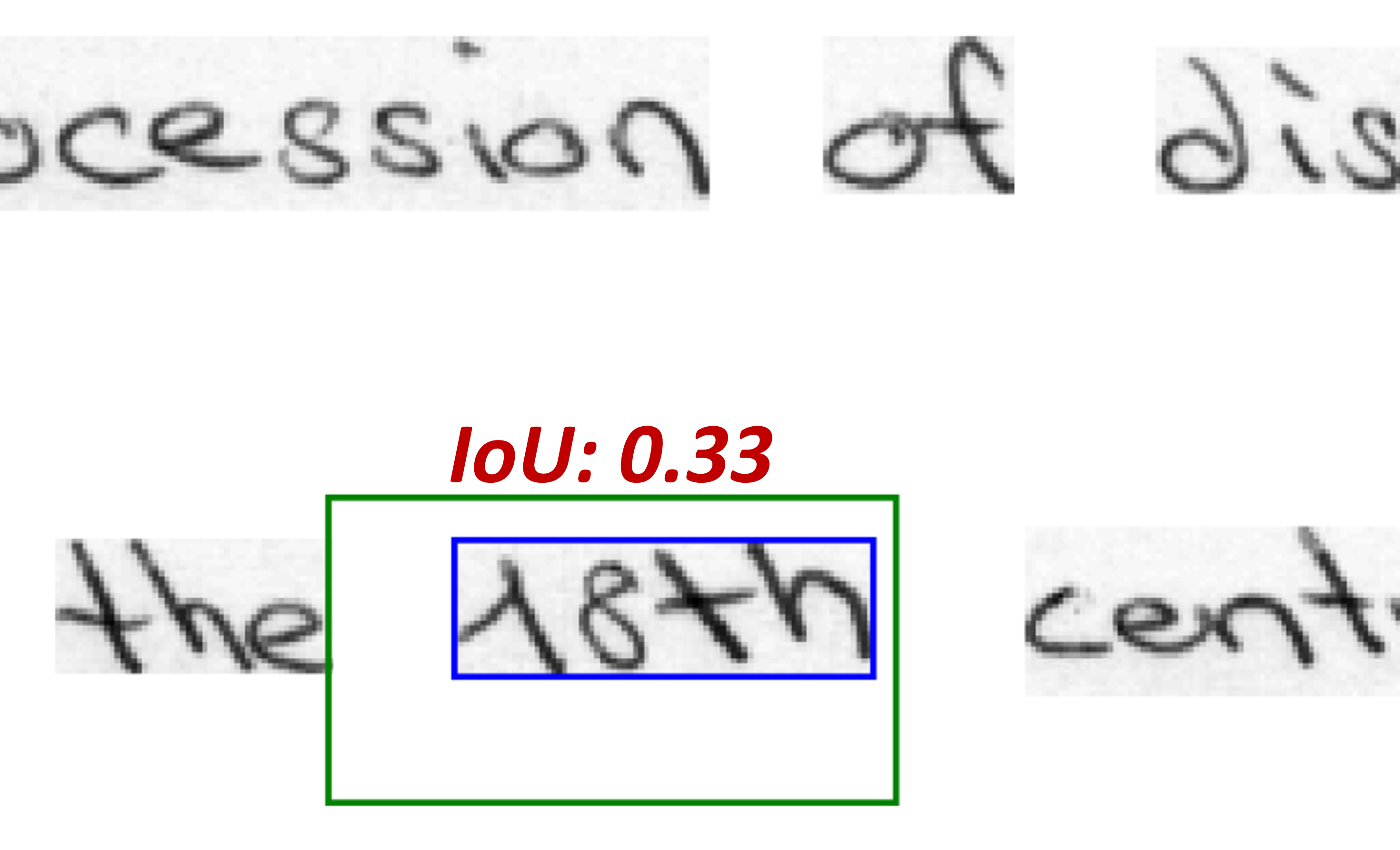}   &  \includegraphics[width=.38\linewidth]{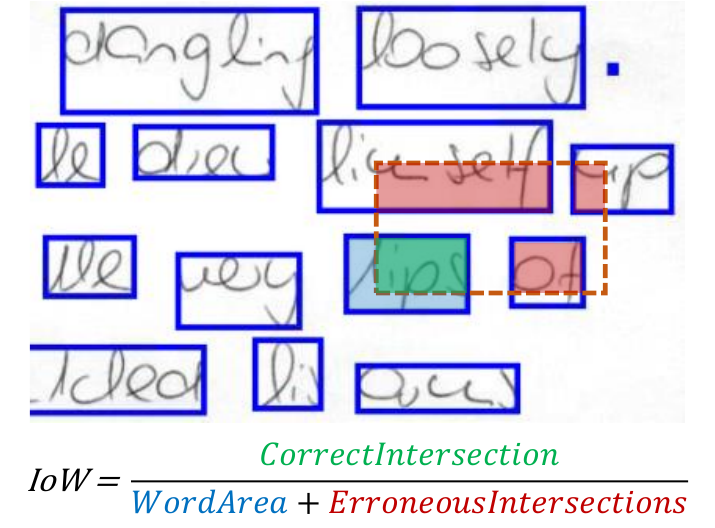}\\
    \end{tabular}
    \caption{Visualization of the IoU issue (left), where a good detection is overly penalized, and of the proposed IoW metric (right), where blue shaded is the original gt area, green shaded the overlap with the gt word and red shaded the erroneous overlaps.}
    \label{fig:iow}
\end{figure}

Objection detection is typically evaluated with mean Average Precision (mAP) if the detected bounding box 
considerably overlaps with the corresponding gt box. 
This overlap is quantified through the Intersection over Union (IoU) metric.
Although the usefulness of such an overlap metric is indisputable, it may not be the most suitable for the word spotting application. 
Specifically, in documents, we have disjoint entities of words that do not overlap. 
A good detection could be defined as one that includes the word of interest and no other (neighboring) word.
However, the IoU metric sets a relatively strict constraint of how the boxes interact. 
For example, in Figure~\ref{fig:iow}(left) we have an arguably good detection of the word, denoted by a green box, that has a very low IoU score of 0.33 with respect to the groundtruth blue box.
This phenomenon is frequent when using our method, since we have not imposed constraints of how this box should be, only that it should contain the word that we are interested in. 
Therefore, for lower IoU thresholds we report very good performance that deteriorates quickly when increasing the threshold, even though the detections are notably spot-on.

To address this issue, we propose a different metric that does not penalize enlarged boxes as long as no other neighboring word is intersected by the detected box.
Specifically, we want to penalize such erroneous intersections and we include their total area into the denominator of our new metric, 
along with the area of the groundtruth box. 
The numerator, as usual, is the intersection of the detected box with the groundtruth box. 
This concept is clearly visualized in Figure~\ref{fig:iow}(right).

\section{Experimental Evaluation}

\textbf{Datasets:}
We evaluated the proposed system in two widely-used English datasets. The first one 
is the challenging \textbf{IAM} dataset~\cite{marti2002iam},  
consisted of handwritten text from $657$ different writers and partitioned into writer-independent train/val./test sets. 
IAM grountruth has indication of erroneous word-level annotation that was used to mask out these regions from the page-level images. 
The second dataset is a collection of manuscripts from George Washington (GW)~\cite{fischer2012lexicon}, consisted of 20 pages with 4860 words in total. 
The split into training/test sets follows the protocol of \cite{wilkinson2017neural} with a 4-fold validation scheme. 

Following the standard paradigm for these two datasets, we ignore punctuation marks and we merge lowercase and uppercase characters. All possible words that do not contain any non-alphanumeric character are considered as queries. Only for the case of IAM, queries that belong to the official 
stopword list are removed from the query list~\cite{Almazan14PAMI, sudholt2016phocnet}.

\noindent \textbf{Metrics:}
We evaluate our approach using the standard metric for retrieval problems, Mean Average Precision (MAP). 
Since the task at hand is segmentation-free spotting, we also utilize a overlap metric between detections and ground-truth boxes.
Following the discussion of Sec.~\ref{sec:iou} we consider three different overlap metrics: 
\emph{1)} standard Intersection over Union (IoU) \emph{2)} a modified IoU, dubbed as x-IoU, that focuses on the overlap over the x-axis by assuming the same y-coordinates between the detection and the groundtruth, while requires an initial overlap of over 0.1 IoU. This resembles a more line-focused metric.\emph{3)} the proposed IoW metric of Sec.~\ref{sec:iou}.

\noindent \textbf{Implementation Details:} 
Training of our model was performed in a single NVidia 1080Ti GPU using the Pytorch framework.
We trained our model for $80$ epochs with Adam optimizer~\cite{kingma2014adam} along with a cosine annealing scheduler, 
where learning rate started at $1e-3$. 
The proposed spotting method, applied over the feature map extracted by the CNN, is implemented with cpu-based Numba~\cite{lam2015numba} functions in order to achieve efficient running time. 
GPU-based implementation of such actions has not been explored, since specific operations cannot be straightforwardly ported with efficient Pytorch functions (e.g., binary search). 
Nonetheless, implementation on GPU could considerably improve time requirements as a potential future extension.
Code is publicly available at \url{https://github.com/georgeretsi/SegFreeKWS}.


\subsection{Ablation Studies}

Every ablation is performed over the validation set of IAM using a network of $\sim6M$ parameters. The typical IoU threshold is used, unless stated otherwise. 

\noindent \textbf{Impact of Spotting Modifications:} 
Here, we will explore the impact of hyper-parameters (thresholds) selected in the proposed spotting algorithm. 
Specifically, we focus on discovering the sensitivity of the candidate pruning thresholds, since this is the only step that introduces critical hyper-parameters.  
To this end, we perform a grid search over the probability threshold $p_{thres}$ and the centering ratio threshold $r_{thres}$, 
as reported in Tables~\ref{tab:hpar_cos}. 
Only the simple cosine similarity of the character count histogram has used in this experiment and thus the MAP scores are relatively low. 
As we can see, the $0.5$ ratio threshold to be the obvious choice, while, concerning the probability threshold, both $0.01$ and $0.05$ values provide superior performance with minor differences and thus $0.05$ value is selected as the default option for the rest of the paper, since it provides non-trivially faster retrieval times.

\begin{table}[!ht]
    \centering
    \begin{tabular}{|c|c|c|c|c|c|}
    \hline
        \backslashbox{$p_{thres}$}{$r_{thres}$} & 0.25 & 0.33 & 0.5 & 0.66 & 0.75  \\ \hline
        0.001  & 65.85 (2.422)  & 65.96 (2.209)  & 66.15 (1.243)  & 54.66 (0.499)  & 35.11 (0.363)  \\
        0.01 & 67.57 (1.101)  & 67.59 (1.006)  & 67.97 (0.628)  & 53.30 (0.317)  & 32.54 (0.234) \\ 
        0.05 & 67.82 (0.643)  & 67.85 (0.624)  & \textbf{67.99} (0.433)  & 52.02 (0.263)  & 31.17 (0.224)  \\ 
        0.1 & 67.53 (0.569)  & 67.55 (0.535)  & 67.67 (0.389)  & 51.49 (0.253)  & 30.82 (0.222) \\ 
        0.2 & 67.09 (0.509)  & 67.10 (0.475)  & 67.16 (0.355)  & 51.00 (0.249)  & 30.47 (0.215)  \\ 
        0.5 & 66.03 (0.445)  & 66.04 (0.428)  & 66.04 (0.322)  & 49.38 (0.238)  & 29.66 (0.203)  \\ 
        \hline
    \end{tabular}
    \vspace{.1cm}
        \caption{Exploration of pruning thresholds with the character counting retrieval method - only cosine similarity over character histograms was used. We report MAP \@ 25\% IoU Overlap \& time (in parenthesis) to retrieve bboxes for a (image, query) pair. }
    \label{tab:hpar_cos}
\end{table}

\noindent\textbf{Impact of Scoring Methods:}
As we can see from the previous exploration, relying only on character counting leads to underperformance. Here, we will explore the impact of the different scoring methods proposed in Section~\ref{sec:scoring}.
We distinguish different strategies according to the use of the CTC re-scoring step. Specifically we report results without the re-scoring step and with the one-way (adjusting the bound on the right of the box) or two-way (adjusting both horizontal bounds using a reverse pass over the sequence - see Sec.~\ref{sec:scoring}) CTC step. We also report results for the multilevel pyramidal representations (denoted as PCount) of character counting and the PHOC alternative, implemented as a thresholded version of the counting histogram that does not exceed 1 at each bin.
These results are summarized in Table~\ref{tab:scoring}, where we also report the time needed for a query/image pair. 
The following observation can be made: \\
$\bullet$ As expected, adding the CTC scoring step, the results are considerably improved. The two-way CTC score approach achieves the best results overall, regardless the initial step (e.g., \# levels, PHOC/PCount). \\
$\bullet$ PCount provides more accurate detections compared to PHOC when a single level is used, but this is not the case in many configurations for extra levels. \\
$\bullet$ For the CTC score variant, we do not see any improvement when using more level on the first step. This can be attributed to the fact that extended box proposals was a common error, as shown in Fig.~\ref{fig:ambig}(right), and thus sub-partitioning of the box does not correspond to actual word partition. \\
$\bullet$ Time requirements are increasing as we use more levels. Furthermore, as expected, the increase when using the CTC score approach and especially the two-way variant. Nonetheless, due to its performance superiority, we select the 1-level PCount first-step along with the two-way CTC score as the default option for the rest of the paper.

\begin{table}[h]
    \centering
    \begin{tabular}{|c|P{1.1cm}|P{1.1cm}|P{1.1cm}||P{1.1cm}|P{1.1cm}|P{1.1cm}||P{1.1cm}|P{1.1cm}|P{1.1cm}|}
    \hline
    & \multicolumn{3}{c||}{w/o ctc-score} & \multicolumn{3}{c||}{w/ one-way ctc-score} & \multicolumn{3}{c|}{w/ two-way ctc-score} \\
    \hline
    levels & PHOC & PCount & time & PHOC & PCount & time & PHOC & PCount & time\\
    \hline
     1 & 64.60 & 67.99 & 0.44 & 85.29 & 85.58 & 0.68 & 88.90 & \textbf{88.98} & 0.81 \\
     2 & 72.57 & 72.24 & 0.50 & 86.30 & 85.83 & 0.69 & 88.90 & 88.88 & 0.82\\
     3 & 73.45 & 72.65 & 0.62 & 86.58 & 86.06 & 0.77 & 88.74 & 88.55 & 0.92\\
     4 & 72.76 & 71.52 & 0.78 & 86.31 & 86.11 & 0.87 & 88.24 & 88.18 & 0.98\\
     5 & 71.14 & 69.82 & 0.93 & 85.86 & 85.60 & 1.01 & 87.85 & 87.61 & 1.09\\
     \hline
    \end{tabular}
    \vspace{.1cm}
    \caption{Impact of different scoring approaches: PCount vs PHOC for different levels / use of CTC scoring step (one-way or two-way) .We reported MAP \@ 25\% IoU Overlap \& time (in miliseconds) to retrieve bboxes for a (image, query) pair. }
    \label{tab:scoring}
\end{table}

\begin{figure}[h]
\vspace{-.2cm}
    \centering
    \includegraphics[width=.65\linewidth]{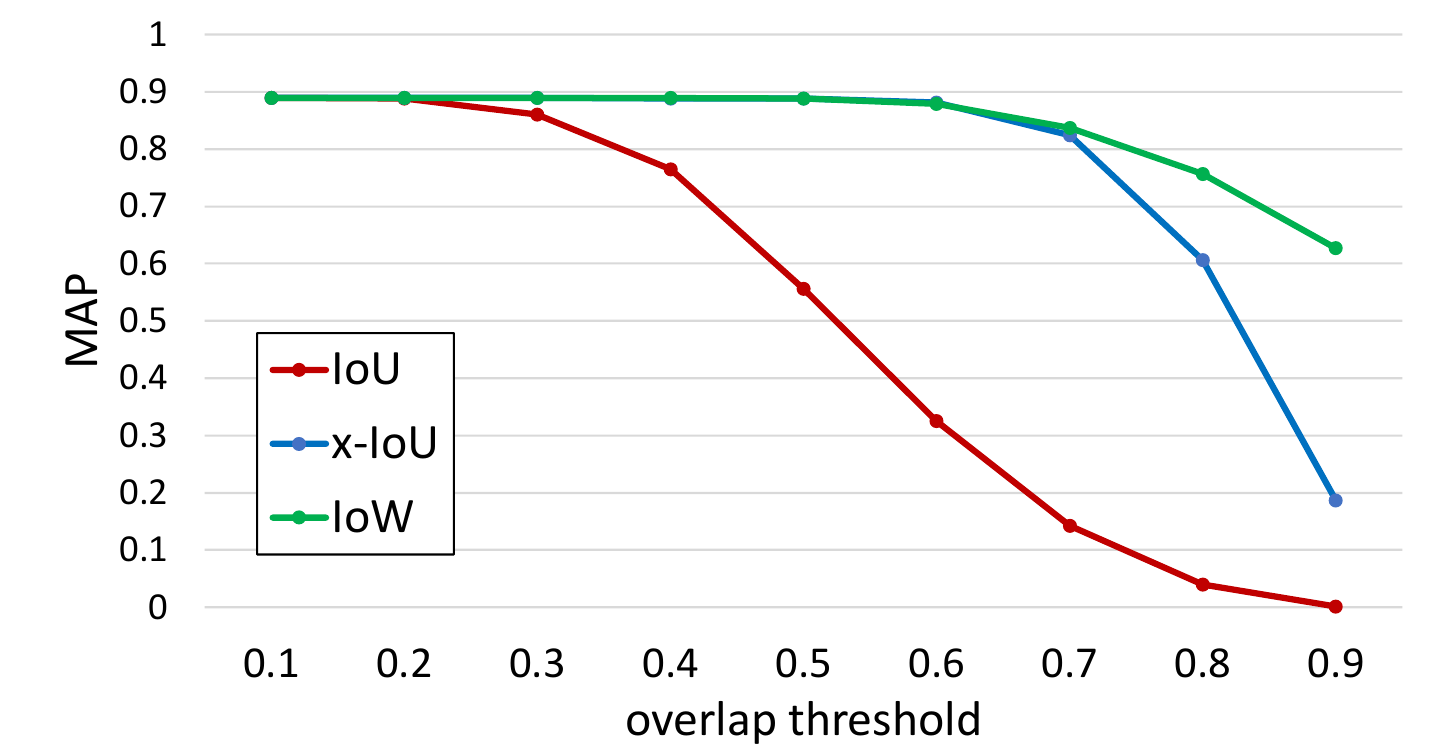}
    \vspace{-.2cm}
    \caption{MAP performance for different overlap thesholds over 3 different overlap metrics: IoU, x-IoU and IoW.}
    \label{fig:iou}
    \vspace{-.5cm}
\end{figure}

\noindent \textbf{Comparison of Overlap Metrics:}
Even though we presented notable MAP results for the $25\%$ IoU overlap, we noticed that the performance considerably decreases as the overlap threshold increases. Specifically, for our best-performing system, the MAP drops from $88.98\%$ to only $54.68\%$.
Preliminary error analysis showed that the main reason was the strict definition of IoU for word recognition, as described in Sec.~\ref{sec:iou}, where many correct detections presented a low IoU metric.
To support this claim, we devised two different overlap metrics (x-IoU and IoW) and we validated the attained MAP for a large range of threshold values, as shown in Figure~\ref{fig:iou}.
As we can see, both the alternatives provide almost the same performance up to $60\%$ overlap threshold whereas rapidly decreases from early on. 
Overall the proposed IoW metric has very robust performance, proving that the main source of performance decrease for larger overlap thresholds was the ``strict" definition of what a good detection is for word spotting applications (see Sec.~\ref{sec:iou} for details).

\newpage

\subsection{State-of-the-Art Comparisons}

Having explored the different ``flavors" of our proposed systems, we proceed to evaluate our best-performing approach (single-level counting similarity along with two-way CTC score) against state-of-the-art segmentation-free approaches.  
Apart form evaluating over the test set of IAM, for the GW dataset, we follow the evaluation procedure used in~\cite{rothacker2015segmentation, wilkinson2017neural} where two different 4-fold cross validation settings are considered. The first assumes a train set of 15 pages, setting aside 5 for testing, while the second assumes a more challenging 5/15 split for 
training/testing, respectively. In each setting, the reported value is the average across the 4-fold setup. We trained our models only on the pages available and not pre-training was performed.
The results are summarized in Table~\ref{tab:sota}, where MAP values for both $25\%$ and $50\%$ overlap thresholds are reported.
The presented results lead to several interesting observations:\\
$\bullet$ the proposed method outperforms substantially the compared methods (even the complex end-to-end architecture of ~\cite{wilkinson2017neural}), for the challenging cases of GW5-15 and IAM, when the $25\%$ IoU overlap is used.
In fact, the proposed method is robust even when using limited training data (e.g., GW5-15), since it does not learn both detection and recognition tasks, but only relies on recognition. \\
$\bullet$ 
As we described before, performance is decreased for the $50\%$ IoU overlap, but the different overlap metrics (x-IoU and IoW) show the effectiveness of the method for larger overlap thresholds, as shown in Table~\ref{tab:sota}(b). Nonetheless, this is not directly comparable with the other compared methods that relied on IoU. \\
$\bullet$ For the case of GW15-5, we do not report results on par with the SOTA, even though the same method has a considerable boost for GW5-15, where fewer data were used. Error analysis showed that for GW, proposals were extended in the y-axis also, resulting to low overlap scores. In other words, the "rough" localization of our method leads to this reduced performance (also discussed in limitations).

\begin{table}[h]
\vspace{-.2cm}
    \centering
    \begin{tabular}{|l||P{1.2cm}|P{1.2cm}||P{1.2cm}|P{1.2cm}||P{1.2cm}|P{1.2cm}|}
    \hline
    & \multicolumn{2}{c||}{GW 15-5} & \multicolumn{2}{c||}{GW 5-15}  & \multicolumn{2}{c|}{IAM} \\
    \hline
    method & 25\% & 50\% & 25\% & 50\% & 25\% & 50\%\\
    \hline
    BoF HMMs \cite{rothacker2015segmentation} & 80.1 & 76.5 & 58.1 & 54.6 & - & - \\
    BG index~\cite{ghosh2015query} & - & - & - & - & - & 48.6 \\
    Word-Hypothesis~\cite{rothacker2017word} & 90.6 & 84.6 & - & - & - & - \\
    Ctrl-F-Net DCToW~\cite{wilkinson2017neural} & 95.2 & 91.0 & 76.8 & 73.8 & 82.5 & 80.3 \\
    Ctrl-F-Net PHOC~\cite{wilkinson2017neural} & 93.9 & 90.1  & 68.2 & 65.6 &  80.8 & 78.8 \\
    Resnet50 + FPN~\cite{zhao2020query} &  96.5 & 94.1 & - & - & - & - \\
    \hline
    Proposed (IoU) & 91.6 & 66.4
 & 85.9 & 66.3 & 85.8 & 59.2\\
    \hline
    \end{tabular} 
    \vspace{.1cm}
    \\
    (a)
    \\
    \vspace{.1cm}
    \begin{tabular}{|l||P{1.2cm}|P{1.2cm}||P{1.2cm}|P{1.2cm}||P{1.2cm}|P{1.2cm}|}
    \hline
    & \multicolumn{2}{c||}{GW 15-5} & \multicolumn{2}{c||}{GW 5-15}  & \multicolumn{2}{c|}{IAM} \\
    \hline
    Proposed (IoU) & 91.6 & 66.4
 & 85.9 & 66.3 & 85.8 & 59.2\\
    \hline
    Proposed (x-IoU) & 93.2 &	92.9 & 
 86.8 & 86.7 & 86.9 & 	86.8\\
    Proposed (IoW) & 92.7	&  87.6
 & 86.8 &	83.0 & 86.9 & 	86.3\\
    \hline
    \end{tabular}\\
    \vspace{.1cm}
    (b)
    \\
    \vspace{.1cm}
    \caption{ (a) MAP comparison of state-of-the-art approaches for IAM and two variations of GW dataset. Both $25\%$ and $50\%$ overlap thresholds are reported. (b) We also report the performance of our method when the proposed overlap metric alternatives are considered (x-IoU, IoW).}
    \label{tab:sota}
\end{table}

\begin{figure}[t]
    \centering
    \begin{tabular}{c | c}
    query & top-6 retrieved boxes\\
    \hline 
    1775 &\raisebox{-.25\height}{ \includegraphics[ width=.72\linewidth]{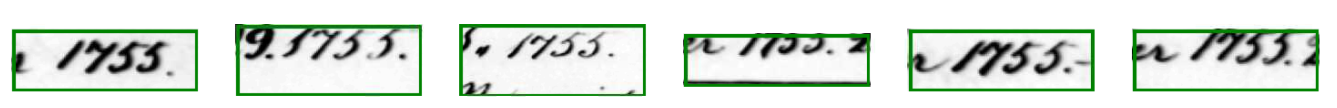}} \\
    soon & \raisebox{-.25\height}{\includegraphics[width=.72\linewidth]{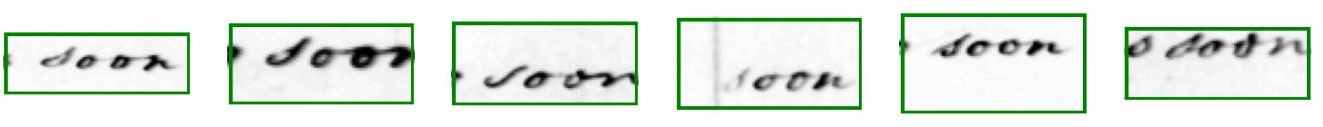}} \\
    them & \raisebox{-.25\height}{\includegraphics[width=.72\linewidth]{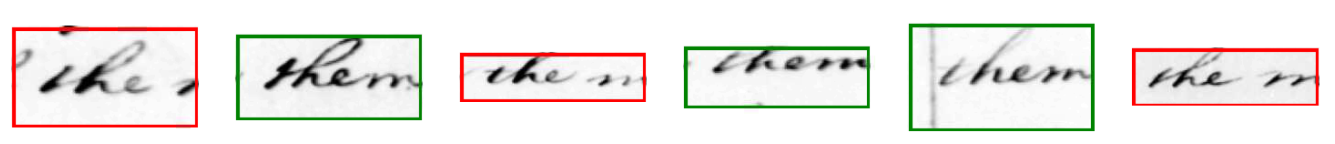}} \\
\end{tabular}
\caption{Examples of QbS with top-6 retrieved words reported for the GW dataset. Green box corresponds to an overlap greater than $25\%$ IoU, while red to lower.}
    \label{fig:retrieval_examples}
\end{figure}

\begin{figure}[t]
    \centering
    \begin{tabular}{cc}
    \includegraphics[width=.48\linewidth]{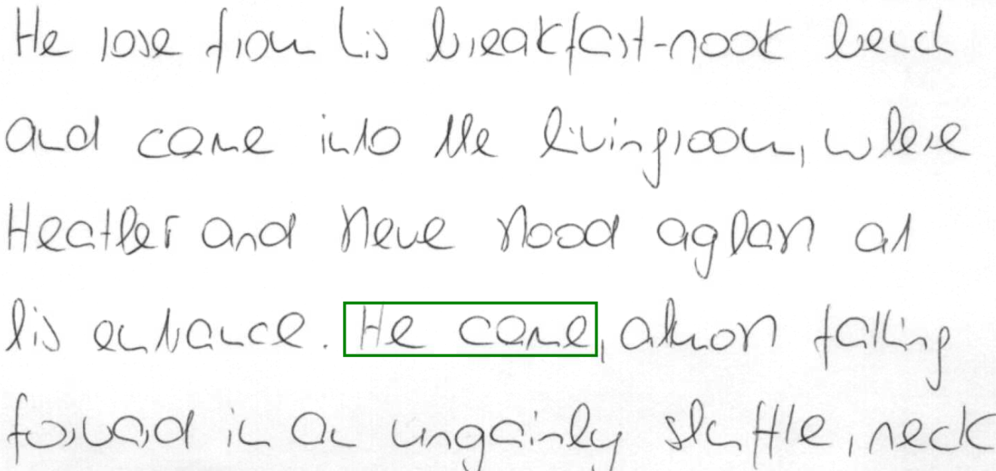}
    &
    \includegraphics[width=.48\linewidth]{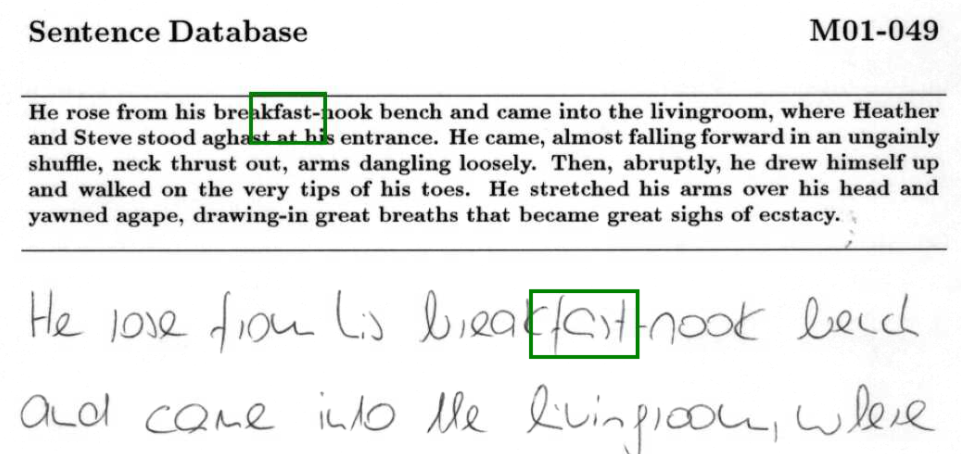} \\
     (a) Query: ``he came"    & (b) Query: ``fast" \\
    \end{tabular}
    \caption{Examples of multi-word and sub-word detections.}
    \label{fig:visual_examples}
\end{figure}

\noindent \textbf{Visual Examples:}
Figure~\ref{fig:retrieval_examples} contains examples of retrieval for the GW dataset, where we can see that retrieved boxes are not tight and can be extended. Specifically, for the case of the query 'them', we retrieved erroneous boxes where the word 'the' appears followed by a word that starts with 'm'. 
In Figure~\ref{fig:visual_examples}, we present some examples of successful multi-word and sub-word retrieval in the IAM dataset. Notably, in the sub-word case of the``fast" query, the fast suffix was detected for the word breakfast in both the handwritten text and also the typewritten reference text of different scale at the top of the image. 

\noindent \textbf{Limitations:} 
$\bullet$ The bounding box proposal stage of Section~\ref{sec:proposed-spotting} tends to provide over-estimations of the actual box, as shown in Fig.~\ref{fig:ambig}(b). This is adequately addressed by the CTC re-scoring step, but such phenomena may cause correct regions to be dismissed before re-scoring. Therefore, a tightened box prediction could improve the overall performance, especially if we assist this with an appropriately designed model component.    
$\bullet$  The proposed approach does not learn to distinguish the space character, as a separator between words, and thus we the ability to detect sub-words can be also be seen as  an issue that affects performance. In fact, if we let sub-words to be counted as correct predictions the MAP (at $25\%$ IoU overlap) increases from $88.98\%$ to $89.62\%$. Even though, detecting sub-words is a desirable property, it would useful if the user could select when this should happen. To add such property, we could include and train the space character as a possible future direction.

\section{Conclusions}
In this work we presented a novel approach to segmentation-free keyword spotting that strives for simplicity and efficiency without sacrificing performance. 
We designed an architecture that enables counting characters at rectangular sub-regions of a document image, whereas it is only trained on single word images. 
The box proposal and scoring steps are designed to speed-up the retrieval of relevant regions, utilizing integral images, binary search and re-ranking of retrieved images using CTC score. The reported results for both GW and IAM dataset prove the effectiveness of our method, while using a simple network of 6M parameters.

\section*{Acknowledgments}
This research has been partially financed by the EU and Greek national funds through the Operational Program Competitiveness, Entrepreneurship and Innovation, under the call ``OPEN INNOVATION IN CULTURE'', project \emph{Bessarion} (T6YB$\Pi$ - 00214).


\bibliographystyle{unsrt}
\bibliography{seg_free_arxiv}

\end{document}